\documentclass[preprint,12pt,nopreprintline]{elsarticle}

\usepackage[T1]{fontenc}
\usepackage[utf8]{inputenc}
\usepackage{lmodern}
\usepackage{amssymb}
\usepackage{amsmath}
\usepackage{booktabs}
\usepackage{tabularx}
\usepackage{adjustbox}
\usepackage{array}
\usepackage{makecell}
\newcolumntype{Y}{>{\raggedright\arraybackslash}X}
\newcolumntype{L}[1]{>{\raggedright\arraybackslash}p{#1}}
\usepackage{graphicx}
\usepackage{url}
\usepackage{hyperref}
\hypersetup{hidelinks}
\begin{document}

\begin{frontmatter}

\title{Revisiting Privacy Preservation in Brain-Computer Interfaces: Conceptual Boundaries, Risk Pathways, and a Protection-Strength Grading Framework}

\author[aff1]{Lei Sun\fnref{eq1}}
\ead{sl20210221@163.com}
\author[aff1]{Xiuqing Mao\fnref{eq1}}
\ead{21166813@qq.com}
\author[aff1]{Shuai Zhang\fnref{eq1}}
\ead{xdzhangs@163.com}
\author[aff1]{Qingyu Zeng\corref{cor1}}
\ead{qingyu6372@163.com}
\author[aff1]{Min Zhao}
\ead{zm717207@163.com}
\author[aff1]{Jiyuan Li}
\ead{erali0314@163.com}
\author[aff1]{Wenle Dong}
\ead{136486145@qq.com}

\fntext[eq1]{These authors contributed equally to this work and should be considered as co-first authors.}
\cortext[cor1]{Corresponding author.}

\affiliation[aff1]{addressline={PLA Information Engineering University}, city={Zhengzhou}, postcode={450001}, state={Henan}, country={China}}

\begin{abstract}
Brain-computer interfaces (BCIs) are moving rapidly from laboratory research into clinical, edge, and real-world settings. Under ISO/IEC 8663:2025, a BCI is a direct communication link between central nervous system activity and external software or hardware systems. This link expands privacy risk beyond raw neural-signal leakage: neural data, derived representations, model assets, and decoded outputs can be re-associated with individuals across collection, transmission, storage, training, inference, and feedback, or used to infer information beyond what a task requires. Starting from the general BCI paradigm, this review defines privacy-protection boundaries, protection objects, and the relationship between user data privacy and model privacy within a shared risk pathway. It then proposes a three-dimensional framework - protection object, lifecycle stage, and dominant protection-strength level - to classify existing work into four levels of protection strength. Finally, mental privacy and neuroethical risks are treated as open issues, emphasizing that BCI privacy protection should not only obscure data but also disentangle task-irrelevant sensitive information while preserving downstream utility.
\end{abstract}


\end{frontmatter}

\section{Introduction}

\label{sec:introduction}

Brain-computer interfaces (BCIs) are rapidly transitioning from laboratory research to clinical translation, edge deployment, and real-world applications. Invasive BCIs have entered the systematic trial phase for communication restoration, motor control, and neural function reconstruction, while non-invasive BCIs are being extensively explored in scenarios such as rehabilitation training, emotion recognition, sleep monitoring, cognitive load assessment, and consumer-grade interaction \cite{patrickkrueger2024_001,xia2023_004}. Neural data is therefore no longer research material confined to controlled experiments; it has the potential to become a high-value data asset capable of sustainable collection, cross-platform processing, and long-term reuse.

Recent advances in neural decoding, brain signal reconstruction, and generative models have expanded BCI privacy risks from traditional ``data leakage'' to ``inference leakage.'' Non-invasive continuous language decoding has already demonstrated the ability to reconstruct semantic content from brain activity, directly raising concerns about mental privacy \cite{tang2023_092,ienca2017_010}; self-supervised speech decoding further indicates that models can recover perceptual language representations from non-invasive brain recordings \cite{dfossez2023_093}; and empirical progress in low-density EEG image classification and reconstruction suggests that portable, low-threshold brain signal reconstruction is approaching real-world application conditions \cite{guenther2024_094}. Collectively, these advancements demonstrate that BCI privacy protection must not only prevent raw neural-signal leakage, but also prevent neural data, model parameters, decoded outputs, and long-term interaction records from being re-linked to specific identities or used to infer diseases, emotions, cognitive states, preferences, or intentions.

This shift gives BCI privacy issues a sense of urgency distinct from that of general data security. Once neural data is illegally stolen, transferred, or used for unauthorized modeling, the risks extend far beyond the leakage of raw signals. Machine learning security research has demonstrated that attacks such as membership inference, model inversion, gradient inversion, and attribute inference can exploit model parameters, gradient updates, confidence outputs, or interface behavior to reconstruct or approximate training data and its sensitive attributes \cite{shokri2017_015,nasr2019_034,zhu2019_035,geiping2020_036,ganju2018_037,melis2018_038,fredrikson2015_075}. In BCI scenarios, such inferred information may correspond to identity characteristics, disease states, emotional cues, cognitive load, and behavioral intentions; if the system continuously accumulates outputs, templates, and user profiles during long-term interactions, privacy risks may further encroach upon deep ethical boundaries such as mental privacy, cognitive liberty, and personal identity \cite{magee2024_002,recommendation2026_005,szoszkiewicz2025_009,ienca2017_010,yuste2017_011,jwa2022_003}.

BCI privacy protection should no longer be simplified to merely ``preventing EEG data from being seen.'' More accurately, it must address two questions: how can neural data and its derived representations be prevented from being re-linked to specific individuals, and how can this information be prevented from being further used to infer content beyond the boundaries necessary for the task? This risk primarily stems from neural data highly associated with individuals, but it does not end with the data itself. During training, inference, and ongoing deployment, models process, compress, store, and reuse this data; model parameters, gradients, embedding spaces, decision boundaries, and inference interfaces can all become vehicles for privacy risks. Models thus transform privacy risks originally attached to the data into algorithmic risks that can be deployed, invoked, stolen, and exploited.

Existing reviews have summarized major privacy risks and protection techniques in BCI systems \cite{xia2023_004,maiseli2023_022}. Representative studies have further examined federated learning, differential privacy, cryptographic computation, secure aggregation, and EEG-oriented privacy-preserving processing in related neural-data scenarios \cite{mcmahan2017_012,kairouz2021_032,dwork2014_027,near2023_013,gentry2009_029,bonawitz2017_033,popescu2021_014,yan2024_021}. However, these studies are often presented in parallel according to technical approaches or attack types, resulting in three key boundaries that remain insufficiently clear. First, user data privacy and model privacy are often treated as parallel concerns, yet the relationship between them remains unexplained; second, raw signals, intermediate representations, model assets, and decoded outputs are frequently lumped together under the umbrella term ``EEG data,'' making it difficult to determine the actual objects of protection; third, differential privacy, engineering isolation, and cryptographic protection are often placed on the same level of discussion, undermining comparisons of their ability to prevent ``re-identification of individuals from data or models.'' International terminology standards have defined BCI as a systematic communication chain comprising acquisition, processing, decoding, feedback, and external devices \cite{isoiec2025_091}. This suggests that this review should organize the literature primarily around system locations and risk pathways, rather than based on individual data labels.

We develop a unified and operational analytical framework for privacy-preserving BCIs. The contributions are fourfold. First, we conceptualize BCI privacy risk as a pathway that extends from user data privacy to model privacy, with user data privacy as the source and model privacy as an extension arising from parameters, representations, and output behavior after data enters the model. Second, we define core concepts, including user data privacy, model privacy, risk carriers, risk pathways, and protection-strength levels, to provide a shared vocabulary for comparing protection methods. Third, we propose a three-dimensional framework - protection object, lifecycle stage, and dominant protection-strength level - to analyze privacy-preserving methods used during training and inference and to clarify differences in risk pathways and protection strength. Fourth, we discuss mental privacy and neuroethical risks as open issues rather than adding them as a third basic protection object, which keeps the technical taxonomy operational and conceptually bounded.

\section{Definition of the BCI System and Its Neural Mechanisms}

\label{sec:definition-of-the-bci-system-and-its-neu}

\subsection{Standard Definition and Basic Linkage of BCI Systems}

ISO/IEC 8663:2025 defines a brain-computer interface (BCI) in the context of engineering systems as a direct communication link between central nervous system activity and external software and hardware systems, specifying that it can support both brain-to-device control and feedback capabilities or bidirectional communication \cite{isoiec2025_091}. This definition distinguishes BCI from general wearable devices or ordinary human-computer interaction systems: its input is not user-externalized data such as clicks, speech, or behavioral records, but rather a signal chain in which neural activity is converted into commands, states, or feedback signals through sensing, channel acquisition, preprocessing, feature extraction, and neural decoding.

According to the terminology of this standard, a BCI system comprises at least the following components: sensors or transducers, channels, processing units, feature extraction, classification/decoding, and feedback and stimulation. Transducers are responsible for converting brain signals into electrical signals; channels constitute the pathways through which brain signals are acquired, transmitted, and processed; stimulation is used to elicit specific responses; and feedback enables the system to continuously compare its output with the target state and adjust accordingly. A BCI is not a single-point acquisition device, but rather a closed-loop or semi-closed-loop system that links neural activity, engineering-based acquisition, algorithmic interpretation, and external control. Figure 1 summarizes this basic paradigm.

\begin{figure}[htbp]

\centering

\includegraphics[width=0.9\textwidth]{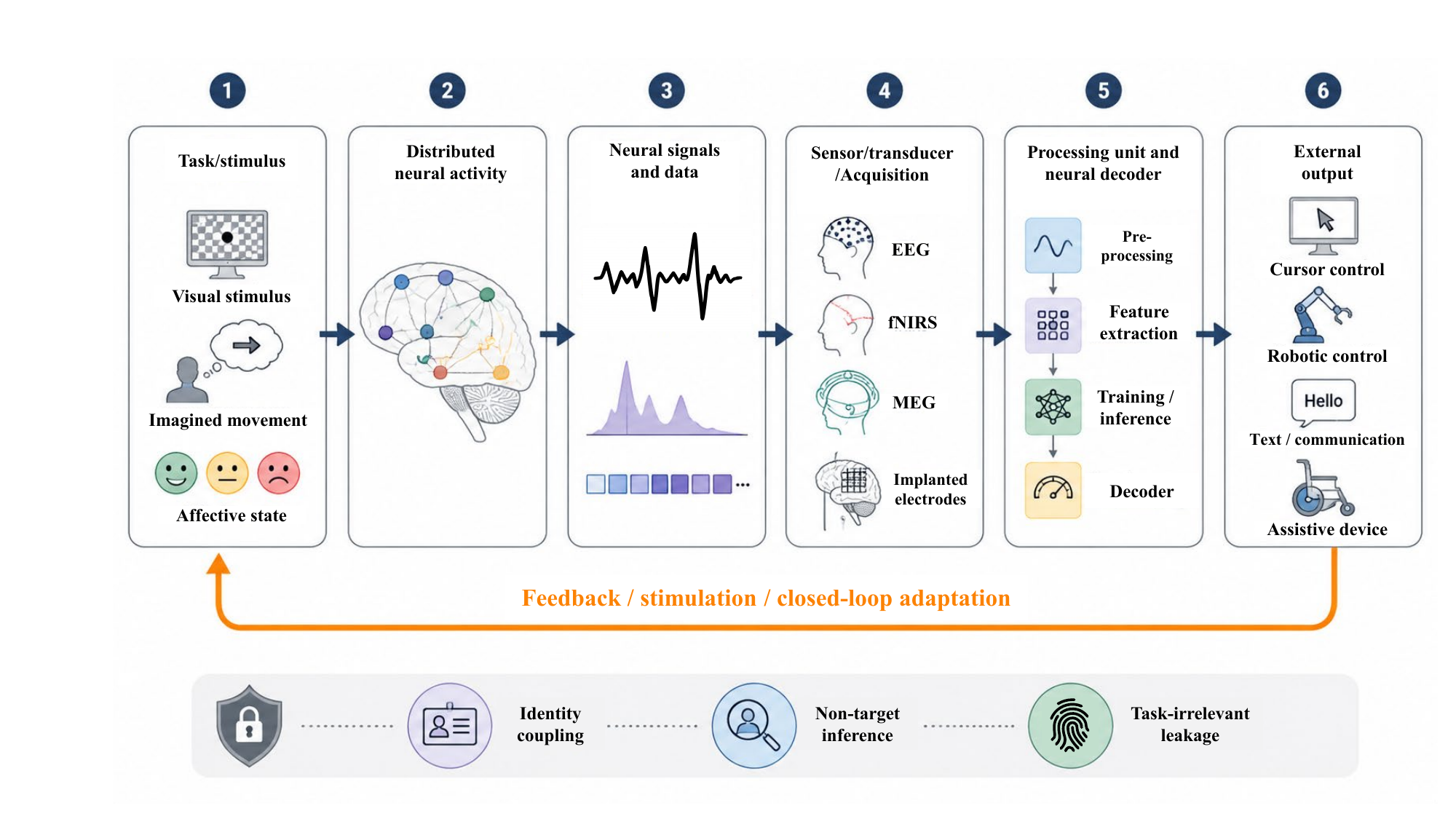}

\caption{General BCI Paradigm and Neural Signal Generation--Decoding Chain}

\label{fig:1}

\end{figure}

The figure illustrates, starting from the task/stimulus or spontaneous regulation, the sequential stages of distributed neural activity, neural signals and data representation, sensor/transducer/channel acquisition, processing units and neural decoding, external output, and feedback loops. The risk zone at the bottom highlights that neural signals may simultaneously carry privacy risks---such as identity coupling, inference of task-irrelevant sensitive attributes, and residual task-irrelevant information---during the processes of generation, representation, and decoding.

\subsection{General Paradigms: Stimuli, Tasks, Neural Encoding, and Decoding}

From a paradigmatic perspective, BCIs can be categorized as active, reactive, affective, synchronous, asynchronous, unidirectional, or bidirectional. A common feature across these paradigms is that the system must induce observable neural activity through tasks, stimuli, or spontaneous regulation, and then interpret this activity as intent, state, category, or control commands through signal processing and neural decoding. ISO/IEC 8663:2025 summarizes signal processing as the transformation, extraction, and classification of brain signals to interpret user intent and control external devices; neural encoding emphasizes the integration and encoding of information by neurons, while neural decoding interprets the electrical signals generated by brain neurons as meaningful information \cite{isoiec2025_091}.

This process determines that BCI privacy risks cannot be understood solely in terms of ``whether data is visible.'' Neural signals are not isolated fields but learnable representations formed through the combined effects of specific stimuli, tasks, brain regions, time windows, and individual physiological states. Signal fragments used for motor imagery classification may simultaneously carry information regarding identity, fatigue, attention, disease status, or emotional differences; features used for emotion recognition may also retain spatial distributions and spectral patterns sufficient to re-identify an individual. Observable data are therefore often the result of integration across multiple brain regions, temporal scales, and cognitive processes \cite{buzski2004_100,friston2011_101}.

\subsection{The Distinctive Nature of Neural Data, Neural Signals, and BCI Privacy}

Here, ``neural signals'' refer to time series or response patterns generated by neural activity and captured via electrical, magnetic, optical, or blood oxygenation methods; ``neural data'' includes raw neural signals, intermediate representations obtained through preprocessing or feature extraction, labels, templates, model inputs and outputs, interaction logs, and long-term derived profiles \cite{yang2025_006,yuste2023_023,isoiec2025_091}.

The distinction is that neural signals emphasize observable activities originating from neural mechanisms, whereas neural data emphasizes the forms of storage, processing, and reuse of these activities once they enter information systems.

The fundamental difference between BCI and general information systems is that the strong coupling between the human and the data is not accidentally formed through subsequent association analysis, but is pre-embedded by neural mechanisms and signal generation mechanisms. In general information systems, the risk of association can be reduced through de-identification, separate storage, or access control; however, many features in BCI inherently stem from an individual's brain structure, neural response patterns, task strategies, and long-term physiological states. Even if names, identification numbers, or device identifiers are removed, these features may still be re-linked to specific individuals through re-identification, attribute inference, or cross-session association. The fundamental challenge of BCI privacy protection is not only to conceal data, but to suppress the continued coupling of task-irrelevant sensitive information to identity and mental states while simultaneously fulfilling downstream tasks.

\section{Conceptual Foundations and Analytical Framework for Privacy Protection in Brain-Computer Interfaces}

\label{sec:conceptual-foundations-and-analytical-fr}

A systematic account of privacy protection in BCIs must begin with the BCI paradigm itself. A BCI is not a conventional data-acquisition system; it is a closed or semi-closed loop centered on neural activity, stimulation or tasks, channel acquisition, feature extraction, neural decoding, and feedback interaction. Accordingly, this section does not begin by listing sensitive labels. Instead, it addresses four questions: how the paradigm defines risk boundaries, how privacy protection determines the objects to be protected, how the operational workflow generates different data representations, and what level of protection different methods can achieve.

\subsection{Determining Privacy Protection Boundaries Based on BCI Paradigms}

General privacy theories emphasize individual control over the flow of personal information, legitimate processing, and contextual integrity \cite{westin1967_016,nissenbaum2004_017,oecd2026_018,brooks2017_019}, as shown in Table 1. While these theories provide a normative foundation for BCI privacy protection, the scope of protection cannot be determined solely by personal information fields; it must be jointly defined by the mechanisms underlying neural activity and the system's operational paradigm. Active BCI, reactive BCI, affective BCI, and bidirectional BCI collect different signals and have different scopes of task-essential information; once the system retains or learns identity, emotions, cognitive load, or disease status beyond the task objectives, privacy risks shift from data exposure to unauthorized inference. It is therefore necessary to integrate neurotechnology standards, neuroethical norms, and neural decoding research to further define the specific boundaries of BCI privacy protection, as shown in Table 2.

\begin{table}[htbp]
\centering
\scriptsize
\setlength{\tabcolsep}{3pt}
\renewcommand{\arraystretch}{1.35}
\caption{General Privacy Theories and Normative Foundations}
\label{tab:1}
\begin{adjustbox}{max width=\textwidth,max totalheight=0.82\textheight}
\begin{tabularx}{\textwidth}{L{0.15\textwidth} L{0.24\textwidth} L{0.23\textwidth} L{0.30\textwidth}}
\toprule
Source & Core points & \makecell[l]{Understanding of\\Privacy} & \makecell[l]{Implications for BCI\\Privacy} \\
\midrule
Warren \& Brandeis, 1890 \cite{warren1890_102} & Defined privacy as the ``right to be let alone.'' & Private life should be free from arbitrary intrusion. & Early basis for protecting internal mental states in BCI settings. \\

Westin, 1967 \cite{westin1967_016} & Individuals decide when, how, and to what extent information is disclosed. & Privacy centers on informational autonomy and control. & Neural-data collection, sharing, and reuse should remain consent-based and user-controlled. \\

OECD Privacy Guidelines, 1980/2013 \cite{oecd2026_018} & Established governance principles on collection limits, purpose limits, safeguards, transparency, participation, and accountability. & Personal-data processing should satisfy purpose, necessity, and security constraints. & Supports data minimization, purpose limitation, and controlled cross-organizational sharing. \\

Nissenbaum, 2004 \cite{nissenbaum2004_017} & Privacy depends on whether information flows fit contextual roles and norms. & Risk emerges when information flows violate contextual norms. & BCI data become riskier when moved from care or research into consumer use, training, or profiling. \\

ISO/IEC 29100:2011 \cite{isoiec29100_020} & Provides an ICT privacy framework for subjects, controllers, processors, principles, and roles. & Privacy protection requires role-based, accountable, lifecycle-wide controls. & Offers a common vocabulary for assigning responsibilities across users, devices, platforms, and service providers. \\

NIST IR 8062, 2017 \cite{brooks2017_019} & Defines privacy engineering as identifying and managing individual risk from data processing. & Privacy risk stems from identifiable harm caused by processing. & Fits BCI settings where risk comes from inference, linkage, or reverse engineering, not only disclosure. \\
\bottomrule
\end{tabularx}
\end{adjustbox}
\end{table}

\begin{table}[htbp]
\centering
\footnotesize
\setlength{\tabcolsep}{3pt}
\renewcommand{\arraystretch}{1.35}
\caption{Neuroscientific Basis for BCI Privacy Protection Boundaries}
\label{tab:2}
\begin{adjustbox}{max width=\textwidth,max totalheight=0.82\textheight}
\begin{tabularx}{\textwidth}{L{0.15\textwidth} L{0.25\textwidth} L{0.23\textwidth} L{0.29\textwidth}}
\toprule
Type of basis & Representative research or standards & Core content & \makecell[l]{Significance for\\BCI privacy boundaries} \\
\midrule
BCI System Definition & ISO/IEC 8663:2025 \cite{isoiec2025_091} & Defines BCI as a CNS-to-external-system link spanning acquisition, processing, decoding, and feedback. & Protection boundaries should cover the full link, not a single EEG file. \\

Neuroethical Guidelines & Yuste et al., 2017; UNESCO, 2025 \cite{yuste2017_011,recommendation2026_005} & Emphasize privacy, identity, agency, psychological integrity, and freedom of thought. & BCI privacy has a psychoethical dimension beyond ordinary personal-data protection. \\

Language and Semantic Decoding & Tang et al., 2023; D\'{e}fossez et al., 2023 \cite{tang2023_092,dfossez2023_093} & Non-invasive recordings can decode continuous semantic and speech-perceptual representations. & Risk includes inferable mental content, semantics, and intent. \\

Image and Perceptual Reconstruction & Guenther et al., 2024 \cite{guenther2024_094} & Low-density EEG already supports image classification and reconstruction. & Even low-cost or consumer BCIs may enable unintended inference. \\

Mental Privacy and Cognitive Biometrics & Magee, Ienca \& Farahany, 2024 \cite{magee2024_002} & Protection should extend to mental states inferable from neural and behavioral patterns. & Mental privacy is a high-level risk, not just another data field. \\

Lifecycle Governance & Kapitonova et al., 2022; Pauzauskie et al., 2025 \cite{kapitonova2022_020,pauzauskie2025_008} & Neurotechnology expansion requires stronger neural-data governance and cybersecurity-aware privacy frameworks. & BCI protection should span acquisition, training, deployment, sharing, feedback, and reuse. \\
\bottomrule
\end{tabularx}
\end{adjustbox}
\end{table}

As shown in Table 2, the uniqueness of BCI privacy protection is that its protection boundaries are jointly determined by system links and decoding capabilities. BCI involves not only neural signal acquisition but also feature extraction, neural decoding, feedback interaction, and control of external devices. Furthermore, research on neural decoding indicates that brain signals can be used to infer non-explicit information such as semantics, perceptions, identity, and cognitive states. BCI privacy protection cannot be limited solely to personal information fields but must address the system's processing workflow and the risks of overreaching inferences that may arise from it.

In BCI, the boundaries of privacy protection are primarily constrained by ``task necessity.'' Compared to general behavioral data, neural data exhibits stronger individual relevance and inferability: it can be used not only to identify ``who'' but also to infer ``what state a person is in,'' ``what diseases or cognitive characteristics they possess,'' and ``whether specific behavioral intentions exist.'' BCI privacy protection must first decouple sensitive information unrelated to downstream tasks. For example, emotion recognition tasks should not unnecessarily expose identity information; identity authentication tasks should not unnecessarily expose emotions, medical conditions, or cognitive states; and rehabilitation control tasks should not accumulate psychological profiles beyond what is required for motor intent over the long term. On this basis, targeted suppression or protection of this task-external information must be implemented.

\subsection{Definition of Privacy Protection and Scope of Protection}

As used in this document, ``brain-computer interface (BCI) privacy protection'' refers to the use of technical, procedural, and governance mechanisms to restrict unauthorized access to, association of, retrieval of, reverse engineering of, or reuse of neural data, its derived representations, model assets, interface outputs, and long-term derivatives throughout the collection, transmission, storage, training, inference, feedback, and continuous interaction of a BCI system; and to prevent the system from inferring, retaining, or disseminating information regarding an individual's identity, health, emotions, cognitive state, preferences, and intentions beyond the scope of its intended tasks. This definition emphasizes three points: First, the objects of protection extend beyond raw EEG files to include intermediate representations, models, and output derivatives; second, the protection objectives go beyond input confidentiality to include preventing re-identification, attribute inference, model inversion, and task-irrelevant information coupling; third, the level of protection must be determined by considering the lifecycle stage and the prevailing protection strength (PS) level.

\subsubsection{Objects of Protection in the BCI Chain}

Under the above definition, the objects of protection must be identified along the BCI chain. ISO/IEC 8663:2025 defines a BCI as a direct communication link connecting central nervous system activity to external software and hardware systems, and provides unified definitions for terms such as user, subject, channel, processing unit, feature extraction, neural decoding, and feedback \cite{isoiec2025_091}. Accordingly, a typical BCI privacy risk chain can be summarized as follows: the user or subject generates neural activity; channels acquire this activity to form raw signals; the processing unit performs preprocessing and feature extraction; neural decoding or model training then generates parameters, embeddings, interfaces, and outputs; and finally, through feedback and continuous interaction, these elements coalesce into templates, logs, and profiles. Privacy risks do not exist solely at the acquisition end but migrate along this chain between data, models, and outputs. On this basis, this paper distinguishes two primary objects of protection in BCI systems: user data privacy and model privacy. Here, ``user data'' is used in a broad sense to encompass neural signals and associated records generated by or linked to a BCI user or subject; the corresponding privacy implications are further discussed in Sections IV and V.

\subsubsection{Applicable Boundaries of Multimodal Neural Data}

The term ``brain-computer interface'' in the title of this review is not synonymous with EEG. According to the terminological framework of ISO/IEC 8663:2025, BCI focuses on the direct communication link between central nervous system activity and external software and hardware systems; specific channels can originate from various recording methods such as EEG, functional near-infrared spectroscopy (fNIRS), magnetoencephalography (MEG), electrocorticography (ECoG), stereoelectroencephalography (sEEG), cortical/implanted electrodes, and multimodal fusion \cite{isoiec2025_091,naseer2015_095,mellinger2007_096,schalk2011_097}. Existing literature on privacy-preserving technologies primarily focuses on EEG, mainly because EEG is non-invasive, relatively low-cost, wearable, and supported by a large amount of publicly available data, making it easier to conduct reproducible experiments involving federated learning, differential privacy, template protection, and secure inference.

However, in terms of privacy protection targets, different modalities do not alter the basic framework of this review. Regardless of whether signals originate from EEG, fNIRS, MEG, or ECoG/implanted channels, BCI systems must undergo stages such as acquisition, preprocessing/feature extraction, modeling, inference, and feedback, potentially generating four types of data carriers: raw acquired signals, intermediate data representations, model assets, and outputs and long-term derivatives. What truly requires distinction is the intensity of risk and the priority of protection: fNIRS is more readily associated with long-term monitoring of cognitive load, emotions, and rehabilitation status; MEG is commonly used in high-precision scientific and clinical recordings, where data sharing and high-dimensional representation leakage are more prominent;

ECoG, sEEG, and implantable signals are deeply tied to clinical identity, long-term closed-loop control, and high-performance decoding, posing higher risks of identity coupling and continuous feedback; multimodal fusion, in turn, amplifies the risks of cross-source association and profile accumulation. Figure 2 summarizes these differences using a heatmap.

\begin{figure}[htbp]

\centering

\includegraphics[width=0.9\textwidth]{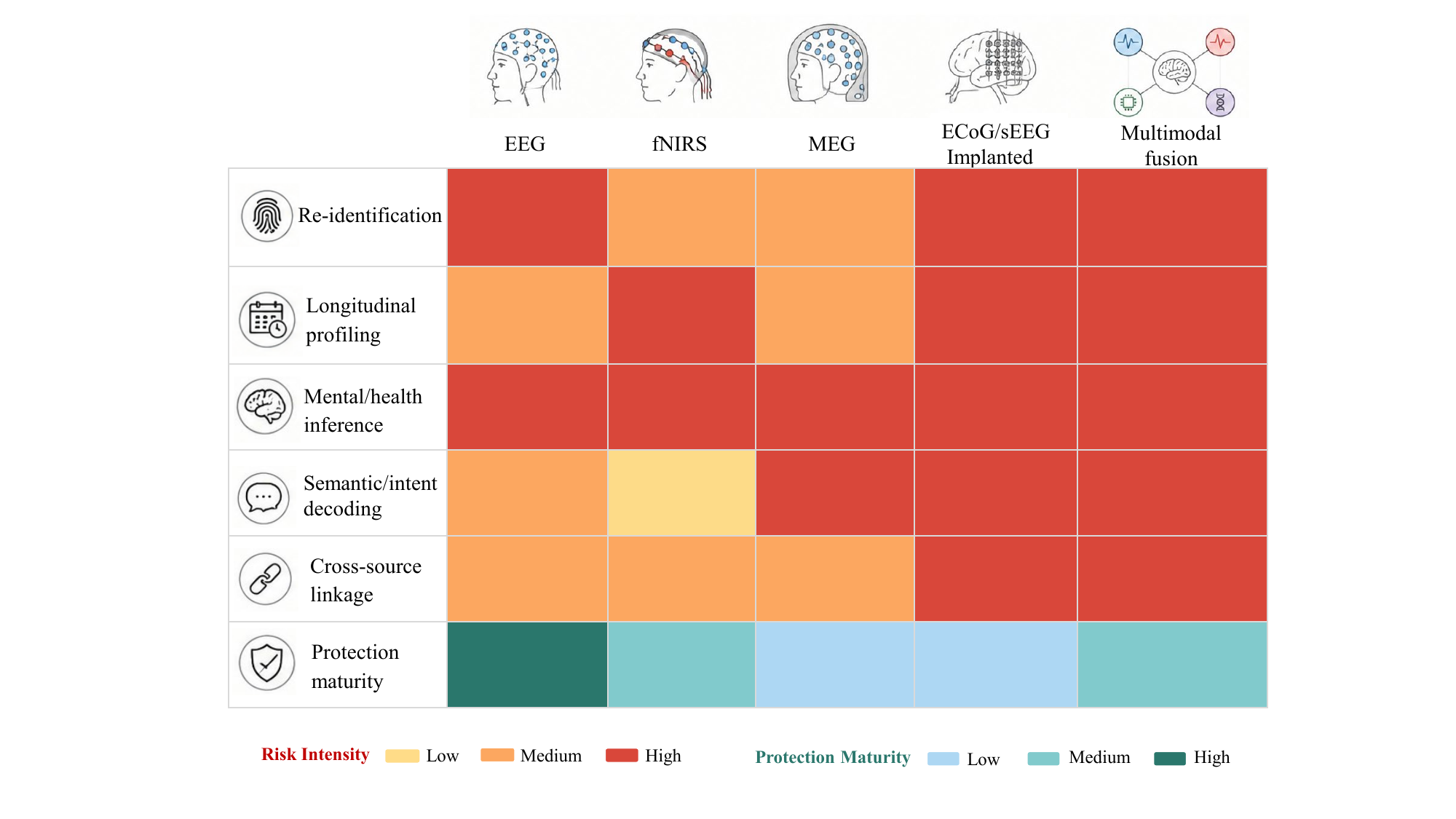}

\caption{Privacy Risk Spectrum Across Different BCI Neural Data Modalities}

\label{fig:2}

\end{figure}

The figure compares the relative differences among EEG, fNIRS, MEG, ECoG/sEEG/implanted signals, and multimodal fusion in terms of re-identification, long-term profiling, state/health inference, semantic/intent decoding, cross-source association, and the maturity of existing protective measures. This figure serves to highlight that while modal differences do not alter the framework presented in this review, they do affect risk intensity, priority protection stages, and the maturity of available methods.

The subsequent methodological review uses EEG as the primary evidence; however, its purpose is not to unconditionally extrapolate EEG conclusions to all BCI modalities, but rather to validate the ``Protection Object---Lifecycle Stage---Dominant Protection-Strength Level'' framework using the most comprehensive technical literature currently available. For fNIRS, MEG, ECoG/implanted signals, and multimodal fusion, this review primarily discusses risk transfer and protection boundaries at the framework level; Figure 2 illustrates that multimodal differences should be understood as variations in risk intensity and applicable boundaries, rather than changes to the analytical framework itself.

\subsection{Data Representations in the Context of the Workflow}

From the perspective of the BCI workflow, neural activity is converted into various data representations once it enters the system. The first category consists of raw acquired signals and preprocessing results specific to individual subjects; the second category includes intermediate data representations such as features, labels, templates, and embeddings; the third category comprises model assets such as parameters, gradients, decision boundaries, and inference interfaces; the fourth category consists of derivatives such as output results, interaction logs, and long-term profiles. The risks associated with these forms vary: the closer they are to the raw data collection end, the more prominent the risks of direct exposure and re-identification; the closer they are to the model and output end, the more prominent the risks of inversion, attribute inference, and long-term profiling.

Model privacy encompasses model parameters, gradients, embedding spaces, decision boundaries, inference interfaces, and other model byproducts. The reason it falls under BCI privacy protection is not because models are inherently equivalent to personal data, but because models may compress, store, and subsequently release the neural features, state distributions, and sensitive attributes of the individuals used for training. Once a model is leaked, replicated, or subjected to long-term querying, it can become a secondary vector for data privacy breaches.

The relationship between the two must be clarified: user data privacy is a source-level issue, while model privacy is an extended issue arising after data enters the model. The consequences of model privacy breaches extend beyond the loss of algorithmic assets; they also negatively impact data privacy, as parameters, embeddings, decision boundaries, and interface outputs may encode individual differences, disease patterns, emotional characteristics, and cognitive states of training samples. In other words, protecting models is not an independent goal detached from data, but a necessary condition for preventing the recovery, transfer, and re-inference of sensitive neural information within algorithmic systems.

\subsection{Risk Carriers and Risk Pathways}

To avoid broadly categorizing all objects as ``EEG data,'' this review classifies the carriers of BCI privacy risks into four categories based on the manifestations of neural data. The first category consists of raw acquired signals, referring to EEG or other neural signals formed after collection by electrodes, sensors, or implantable devices, followed by amplification, front-end filtering, and analog-to-digital conversion; even after hardware amplification and digitization, as long as they have not yet undergone high-level semantic decoding, they can still be regarded as part of the raw signal chain.

The second category is intermediate data representations, which include preprocessing results, segmented signals, time-frequency features, spatial filtering results, embedding vectors, labels, local representations, and task-related features. ISO/IEC 8663:2025 defines ``feature extraction'' as the process of identifying and isolating relevant signals from brain activity data, and ``neural decoding'' as the process of interpreting electrical signals generated by brain neurons and converting them into meaningful information \cite{isoiec2025_091}. This indicates that intermediate representations are not only low-risk byproducts, but rather a critical transformation layer between signals and intentions, states, and commands.

The third category is model assets, which include gradients, parameters, local updates, aggregated models, model weights, model structures, embedding spaces, decision boundaries, and inference interfaces. While they appear to be algorithmic objects, they may carry traces of training samples, training distributions, or individual attributes, and are therefore the primary carriers of model privacy.

The fourth category consists of outputs and long-term derivatives, including decoding results, control commands, text generation results, identity templates, user profiles, long-term behavioral trajectories, feedback logs, and model memories. These represent the accumulated outcomes of the combined effects of data and models; they are often the most readily reused by the system over time and are most likely to accumulate into stable profiles of individuals through long-term interactions.

The four categories described above are not new analytical dimensions, but rather a description of risk carriers: the first two are primarily data-based, the third is model-based, and the fourth consists of the accumulated results of the interaction between data and models. The criteria for distinguishing these categories are not ``whether they have been processed,'' but rather whether they still primarily carry low-level signal structures, whether they have been compressed into trainable representations, whether they carry explicit task semantics, and whether they can accumulate into identifiable or profilable assets through long-term use. The subsequent discussion continues to use the protected subject, lifecycle stage, and dominant protection-strength level as the primary framework.

BCI privacy risks therefore encompass at least five pathways. Exposure refers to unauthorized access to raw signals, features, templates, parameters, or interface information; Association refers to these carriers being re-linked to specific individuals; Inference refers to the system deriving health, emotional, cognitive, or behavioral attributes beyond identity from data or models; Reconstruction refers to restoring input samples or approximating neural features from intermediate representations, gradients, parameters, or outputs; Inversion refers to using model behavior, confidence scores, gradients, or parameters to infer training data or sensitive attributes \cite{shokri2017_015,nasr2019_034,zhu2019_035,geiping2020_036,ganju2018_037,melis2018_038,fredrikson2015_075}.

These pathways illustrate that the challenge in BCI privacy protection does not lie in whether data is ``seen'' at a particular stage, but rather in whether data and models can continue to be used for identifying and interpreting individuals in subsequent processes. Before data enters the model, risks primarily manifest as raw exposure, unauthorized access, and direct association; after data enters the model, risks may transform into membership inference, gradient inversion, model inversion, attribute inference, and reanalysis following model theft. Models do not compete with data for importance; rather, they make data risks persistent, compressed, interface-based, and reusable.

\subsection{Protection Strength Grading and the Three-Dimensional Analysis Framework}

In existing BCI privacy protection research, what truly requires unification is not the sensitivity labels of the protected information, but the protection strength that different methods can achieve in preventing data or models from re-inferring individuals. Accordingly, this review retains the PS1--PS4 (Protection Strength for BCI) numbering and incorporates it into the analytical framework, rather than establishing a separate classification system.

Level 1 Strength (PS1) is the basic exposure control level, which primarily reduces the probability of direct exposure through access control, permission management, local storage, data minimization, interface restrictions, log auditing, basic de-identification, and template substitution. Level 2 (PS2) is the engineering isolation and process constraint level, which primarily restricts visibility within the processing chain through mechanisms such as edge processing, local training, federated learning, separated architectures, trusted execution environments, secure aggregation, module isolation, and runtime access control. Level 3 (PS3) is the statistical perturbation and limited provability level, which includes differential privacy and its variants, perturbed training, controlled output, representation sanitization, and partially revocable representations. Level 4 (PS4) is the cryptographically strong protection level, represented by homomorphic encryption, secure multi-party computation, cryptographic training, cryptographic inference, and strict security protocols. This review classifies differential privacy as PS3, and homomorphic encryption, secure multi-party computation, and cryptographic inference as PS4; PS1--PS4 are not static labels for methods, but rather characterizations of their dominant protection mechanisms and the highest achievable strength in a given scenario.

It should be emphasized that the protection-strength levels serve as a yardstick for comparing the protection mechanisms and blocking capabilities of specific methods in specific scenarios, rather than fixing a particular lifecycle stage to a specific level. Specifically, differential privacy falls under PS3; homomorphic encryption, secure multi-party computation, and cryptographic inference fall under PS4; federated learning, secure aggregation, and trusted execution environments typically correspond to PS2 or approach PS3 in certain aspects; methods such as watermarking, fingerprinting, and template replacement primarily fall within PS1 to PS2, serving more as foundational exposure controls or post-event tracking mechanisms.

Based on the above definitions, this review adopts a three-dimensional analytical framework comprising ``protection object---lifecycle stage---dominant protection-strength level.'' Figure 3 organizes the data carrier locations before and after BCI data enters the model system, the two core protection objects, lifecycle stages, dominant protection-strength levels, and supporting risk pathways within a single framework to serve as a comparative benchmark for the various methods discussed later.

\begin{figure}[htbp]

\centering

\includegraphics[width=0.9\textwidth]{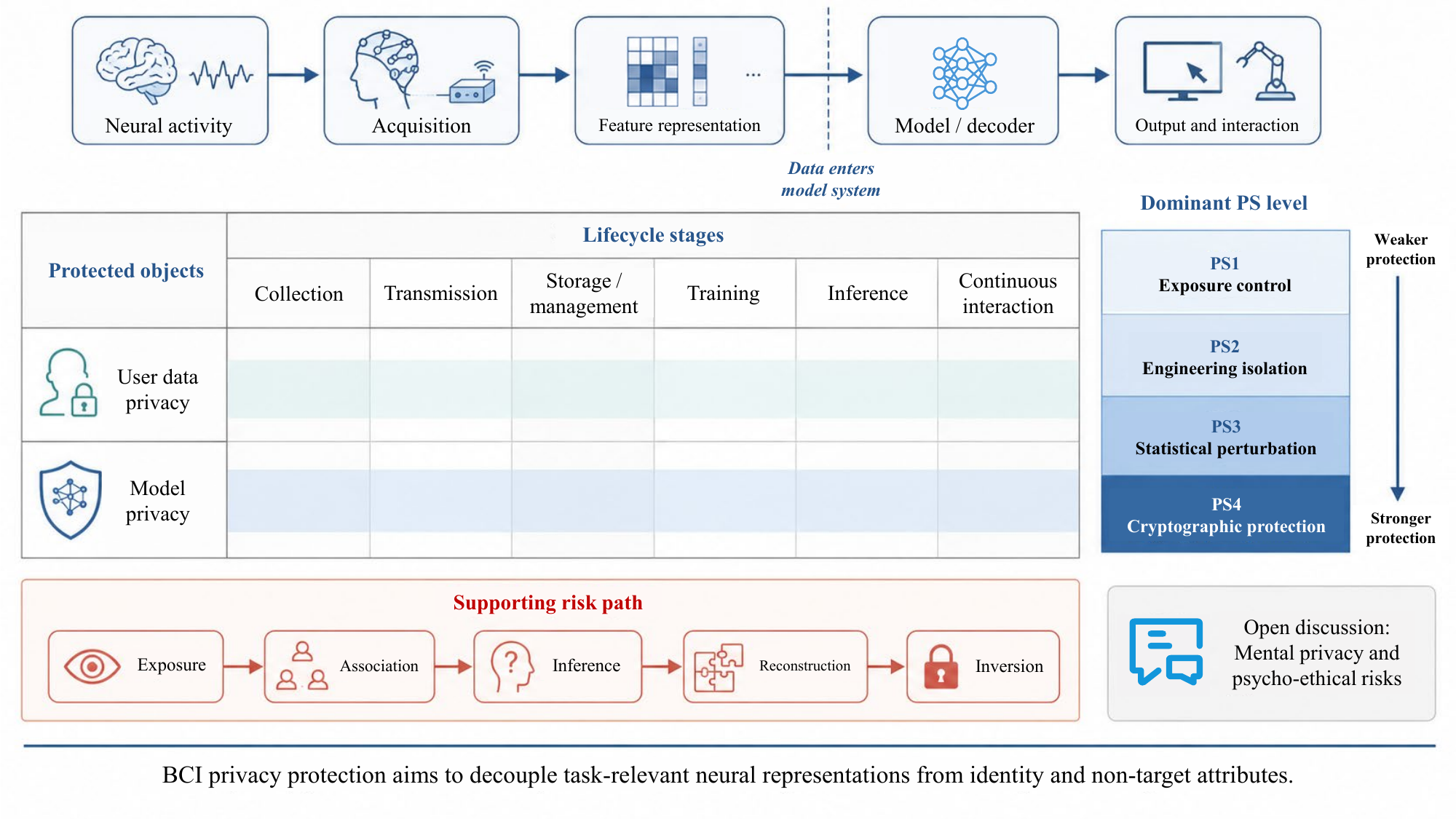}

\caption{Core Analytical Framework for BCI Privacy Protection}

\label{fig:3}

\end{figure}

The upper part of the figure presents a simplified data flow from BCI data to the model. The middle section maps user data privacy and model privacy to lifecycle stages such as collection, transmission, storage/management, training, inference, and continuous interaction. The right side uses PS1--PS4 to indicate the dominant protection intensity of specific methods in particular scenarios. The risk pathways at the bottom illustrate how exposure, association, inference, reconstruction, and inversion support protection-strength level assessments, though they do not constitute independent analytical dimensions; The open-ended discussion in the lower right corner indicates that mental privacy and neuroethical risks are extended governance issues beyond the basic framework.

The first dimension is the objects of protection, which include user data privacy and model privacy. It is important to emphasize that these two objects are not independent starting points at the same level, but rather different points along the same privacy risk chain: data is the source of risk, while the model is the key carrier of risk once it enters the algorithmic system.

The second dimension is the lifecycle stage, encompassing collection, transmission, storage/management, training, inference, and continuous interaction. The training stage focuses on leaks in samples, labels, gradients, parameter updates, and collaborative modeling; the inference stage focuses on leaks in real-time inputs, templates, model interfaces, output feedback, and deployment environments; while the continuous interaction stage addresses cumulative risks arising from long-term logs, user profiling, and model memory.

The third dimension is the dominant protection-strength level, namely the PS1--PS4 classification proposed below. This classification applies to both the control of data visibility and the control of model leakage paths; its evaluation criterion is always the ability to prevent ``re-identifying individuals from data or models,'' rather than the method name itself.

Accordingly, the subsequent sections and tables uniformly address four questions: whether the method protects user data or model assets; at which lifecycle stage the method operates; which type of risk pathway---exposure, association, inference, reconstruction, or inversion---the method primarily blocks; and what the corresponding dominant protection-strength level is. Risk pathways are used to explain protection strength and residual risk and do not constitute a fourth analytical dimension independent of the ``protection object---lifecycle stage---dominant protection-strength level'' framework. The four main research units discussed in this review include: (1) user data privacy during the training phase; (2) user data privacy during the inference phase; (3) model privacy during the training phase; and (4) model privacy during the inference phase.

It should be further clarified that the preceding sections employ the BCI full chain to identify the origins and migration paths of privacy risks; this does not imply that subsequent sections provide an equal overview of all stages, including collection, transmission, storage, training, inference, and feedback. Based on existing literature, research that enables method comparisons and the categorization of privacy levels primarily focuses on the training and inference stages: the former corresponds to leakage risks in multi-center data sharing, parameter updates, representation learning, and collaborative modeling; the latter corresponds to risks related to online inputs, model interfaces, template authentication, secure inference, and output exposure. The collection, transmission, storage, and feedback stages also constitute significant risk entry points, but these are more often manifested as systemic constraints such as data minimization, access control, encrypted transmission, log governance, use restrictions, and ethical compliance; therefore, this review incorporates them into the risk chain and carrier analysis, while the subsequent literature review is primarily focused on the training and inference stages.

\section{User Data Privacy}

\label{sec:user-data-privacy}

Since existing privacy-preserving technical literature primarily focuses on EEG, and comparable protection methods are mainly concentrated in the training and inference stages, the following review centers on EEG and is organized around these two stages; other modalities and processes such as acquisition, transmission, storage, and feedback are primarily used to illustrate framework extrapolation and risk differences, and do not claim to have privacy protection evidence of the same scale. As used in this review, user data privacy refers to the protection of neural data and its derived representations associated with a specific individual during acquisition, transmission, storage, training, and inference, ensuring that such data is not unauthorizedly accessed, correlated, recovered, inverted, or used to infer information beyond what is necessary for the task. The data objects covered here include raw acquired signals, intermediate data representations, templates, labels, logs, and output derivatives. User data privacy is a fundamental issue in BCI privacy risks; its protection objective is not to make data formally ``invisible,'' but to prevent these data carriers from being re-linked to specific individuals or revealing their task-irrelevant sensitive attributes.

\subsection{User Data Privacy in the Training Phase}

User data privacy in the training phase is one of the most focused areas of current privacy-preserving BCI research. The primary threats in this phase include the direct exposure of raw neural data during multi-center sharing; member and attribute inference triggered by parameter exchange and the propagation of intermediate representations; cross-round differential analysis in multi-round collaborative training; and the risk of malicious participants amplifying the theft and manipulation of individual neural features through anomalous updates, backdoors, or poisoning strategies \cite{xia2023_004,shokri2017_015,kairouz2021_032,bonawitz2017_033,nasr2019_034,zhu2019_035,geiping2020_036,ganju2018_037,melis2018_038,zhang2024_039,r2026_040}. In terms of data carriers, this phase primarily involves raw acquired signals, intermediate data representations, labels, local updates, and aggregated parameters; in terms of protection objectives, the core focus is not only on preventing the centralized sharing of raw EEG data but also on limiting information leaks that can be exploited during training interactions. To ensure that the literature review serves the framework of this review, Table 3 summarizes representative methods and the literature context regarding user data privacy in the training phase.

\subsubsection{Federated Learning}

In existing work, federated learning primarily serves the role of ``reducing the flow of raw data sets.'' The basic idea is to keep multi-center EEG or BCI data locally, exchanging only parameter updates or model states, thereby reducing the direct exposure risk associated with centralized sharing \cite{mcmahan2017_012,kairouz2021_032}. In brain-computer interface scenarios, federated learning has been applied to EEG classification, epilepsy detection, and motor imagery tasks, and has mitigated data distribution differences through federated transfer learning or cross-subject collaborative modeling \cite{ju2020_041,amato2025_042}. From the perspective of the PS1--PS4 strength framework, federated learning primarily corresponds to PS2 protection---engineering isolation and process constraints: while it effectively reduces the centralized exposure of raw inputs, it does not automatically impose strict constraints on membership inference, gradient inversion, or attribute inference. Recent work such as SAFE further combines local normalization, adversarial training, and weight perturbation to improve robustness while enhancing collaborative performance under heterogeneous data conditions, indicating that federated learning in brain-computer interfaces should be understood more as an engineering-based collaborative training framework rather than a naturally established high-strength privacy protection mechanism \cite{jia2026_043}.

The value of this class of methods lies in transforming ``centralization of raw neural data'' into ``collaborative model updates,'' making them suitable for multi-center and cross-institutional training; their limitation is that gradients and parameter updates may still carry invertible information, so they typically serve only as procedural isolation rather than a guarantee of full privacy.

\subsubsection{Differential Privacy}

Differential privacy is one of the few typical methods in the training phase that can provide quantifiable statistical leakage constraints. By introducing noise into gradients, parameter updates, or intermediate variables, differential privacy can impose quantifiable constraints on the influence of individual samples on model outputs, thereby reducing the success rate of membership inference and partial attribute inference attacks \cite{dwork2014_027,abadi2016_044}. In the field of brain-computer interfaces, existing research has combined differential privacy with federated learning for tasks such as EEG-based epilepsy recognition and sleep stage classification, and other studies have further explored differential privacy mechanisms in multimodal EEG representation learning \cite{luo2025_045,yin2022_046,fu2024_047}. According to the classification in this review, differential privacy belongs to the PS3 statistical perturbation and restricted provability class: it possesses formalized privacy budgets and statistical bounds, but does not equate to blocking plaintext access or eliminating all semantic inference risks. BCI data is typically high-dimensional, has a low signal-to-noise ratio, and exhibits strong inter-individual variability; introducing noise often more readily compromises discriminative information. Consequently, its engineering feasibility is constrained by significant trade-offs between privacy budgets, accuracy loss, and convergence stability.

Its key value lies in providing comparable privacy budgets; however, in BCI scenarios characterized by small sample sizes, large inter-subject variability, and low signal-to-noise ratios, the conflict between noise intensity and task accuracy becomes more pronounced.

\subsubsection{Secure Aggregation}

Secure aggregation further provides protection against ``direct visibility in parameter exchange.'' The protocol proposed by Bonawitz et al. ensures that the server receives only the aggregated results of model updates, without access to the raw gradients or parameter updates of individual clients \cite{bonawitz2017_033}. From the perspective of the PS1--PS4 strength framework, such methods are generally better classified under the PS2 level of engineering isolation and process constraints; certain aspects combined with differential privacy or stronger security protocols may approach PS3: their core function is to limit the observable information of specific participants on specific communication links, rather than directly providing a global privacy upper bound. For multi-center collaborative training in brain-computer interfaces, this mechanism offers significant value because it minimizes the risk of individual neural features being directly exposed through single-round updates; however, its effectiveness remains dependent on the adversary's model, system assumptions, and multi-round training conditions, and attackers may still gradually recover individual features through differential analysis of aggregated results \cite{bonawitz2017_033,so2023_050,paul2025_051}. Therefore, secure aggregation functions more as a strong process constraint rather than a definitive, comprehensive privacy guarantee.

\subsubsection{Data Hiding Based on Perturbation, Representation Learning, and Generative Models}

In addition to collaborative training mechanisms, another class of research directly acts on the data or representations themselves, using perturbation, representation sanitization, adversarial representation learning, domain adaptation, and generative modeling to reduce the feasibility of re-identification, cross-session correlation, and inference of sensitive attributes. This approach emphasizes modifying the raw signals or their learnable representations prior to sharing, ensuring that the shared EEG data remains usable for task training while making it difficult to reliably identify individuals or recover other sensitive attributes. Meng et al.'s work on identity-unlearnable EEG and multi-attribute joint privacy protection demonstrates how to significantly weaken the identifiability of attributes such as identity, gender, and BCI experience while preserving main task performance as much as possible \cite{meng2023_052,meng2024_053}; A3E, PAT, domain adaptation methods, and synthetic data generation based on GANs or diffusion models further reduce the risks associated with sharing real neural data at the representation and sample substitution levels \cite{chen2025_054,chen2025_055,huang2024_056,li2022_057,debie2020_058,pascual2021_059,chen2025_060}.

In terms of the prevailing protection-strength levels, this approach generally remains dominated by Level 1 (basic exposure control) and Level 2 (engineering isolation and process constraints). While a few methods can approach Level 3 (statistical perturbation and limited provability) under specific attack scenarios, there is still a lack of a unified formal upper bound for comparison. Its advantages lie in engineering flexibility and the ability to balance utility, while its limitations include protection effectiveness that is highly dependent on attack models and task transfer conditions, and a lack of systematic verification regarding the suppression of deeper psychological semantics.

In this research direction, Meng et al. were the first to propose the representative ``identity-unlearnable'' approach. They pointed out that not only can user identity information be learned from EEG data, but data from different sessions may also be re-linked to the same subject, thereby providing an entry point for subsequent deeper privacy mining. To address this, the work designed two types of perturbations---sample-wise and user-wise---to transform raw EEG into identity-unlearnable EEG data: on one hand, by constraining the consistency of task-related outputs to preserve as much information about the main task as possible; on the other hand, by enhancing the correlation between the perturbations and identity labels, making the model more inclined to learn the perturbation patterns rather than the true identity patterns. Experiments demonstrate that this method reduces the average user identification accuracy by nearly 50\% while maintaining essentially stable task classification performance, effectively mitigating the risks of cross-session association and re-identification \cite{meng2023_052}.

Building on this, Meng et al. further extended the approach to joint protection of multiple privacy attributes. They noted that, in addition to identity, EEG data can reliably leak information such as gender and BCI usage experience; to simultaneously suppress multiple sensitive attributes, the paper generates small perturbations for each privacy attribute and constructs privacy-protected EEG data through superposition, making it difficult for privacy classifiers to recover identity, gender, and experience labels while largely preserving information relevant to the main task. Experimental results show that the classification performance for privacy attributes decreased on average from 62.25\% to 35.70\%, while the average accuracy of the main task remained largely unchanged, achieving a good balance between privacy suppression and task accuracy \cite{meng2024_053}.

In addition to the aforementioned data-level perturbation methods, related studies have also utilized adversarial training, domain adaptation, and generative modeling to construct EEG representations that are insensitive to individual identities or domain differences. For example, methods such as A3E and PAT use adversarial learning to reduce the model's reliance on individual features, thereby enhancing cross-subject generalization capabilities while reducing the risk of identity leakage \cite{chen2025_054,chen2025_055}. Furthermore, domain adaptation and source-independent transfer methods have been employed to reduce the residual privacy information in cross-subject features \cite{huang2024_056,li2022_057}. Concurrently, generative models have been used to construct privacy-preserving synthetic EEG data; for instance, Debie et al. and Pascual et al. utilized GANs to generate EEG signals as substitutes for real data during sharing and training, thereby mitigating the risk of directly exposing raw neural data \cite{debie2020_058,pascual2021_059}.

Building on this foundation, recent research has begun to incorporate diffusion models to further enhance the quality and distributional consistency of generated data. For example, works such as EEGCiD achieve high-fidelity reconstruction and privacy-friendly representations of neural signals by compressing or mapping EEG data into diffusion-based generative frameworks. Compared to traditional GANs, these methods offer advantages in terms of generation stability and distributional fitting capabilities, and are therefore regarded as an important direction for generative methods in brain-computer interface privacy protection \cite{chen2025_060}.

The core of the various methods mentioned in this subsection is closer to pre-training anonymization or pre-sharing transformation: that is, they reduce the feasibility of re-identification, cross-session association, and inference of sensitive attributes by altering the raw data or its learnable representations, rather than providing strict formalized bounds as differential privacy does. Therefore, these methods generally still fall under risk mitigation strategies. At the same time, this approach has several limitations: First, the effectiveness of protection typically relies on transfer relationships between attack models, task models, and perturbation generation models, which may diminish when generalized to different model structures; second, the attributes currently explicitly protected remain relatively limited, primarily focusing on labelable variables such as identity, gender, and experience; third, existing methods lack a unified modeling and systematic evaluation framework for more complex and dynamic privacy semantics, such as emotions, intentions, and cognitive states.

It should also be noted that source protection does not occur solely at the software layer. The space-time-coding metasurface secure wireless BCI scheme proposed by Xiao, Fan, Ma, et al. deeply integrates SSVEP visual stimulus coding with metasurface space-time coding at the physical layer, ensuring that brain signal acquisition, command recognition, and wireless transmission are simultaneously constrained by stimulus frequency, biological response patterns, and harmonic-encoded beams \cite{xiao2025_098}. A key insight from such work is that BCI privacy protection can be shifted to the acquisition and interaction front end: by leveraging human neural response patterns, stimulus coding, and physical-layer encryption, the raw acquisition link gains stronger resistance to eavesdropping and replay attacks before entering conventional machine learning workflows.

\subsubsection{Summary}

When viewed within the three-dimensional framework of ``Protected Object---Lifecycle Stage---Dominant Protection-Strength Level'' and in conjunction with risk pathways, research on user data privacy during the training phase has formed a relatively clear hierarchical structure: Federated learning primarily mitigates the risk of raw dataset sharing in multi-center training, typically corresponding to PS2; differential privacy primarily mitigates risks related to membership inference, attribute inference, and partial gradient inversion, corresponding to PS3; Secure aggregation primarily mitigates the direct visibility of single-client updates during the exchange process, typically corresponding to PS2; when combined with differential privacy or stronger security protocols, specific components can approach PS3; methods such as data-level perturbation, adversarial representation learning, domain adaptation, and generative data substitution primarily mitigate risks related to re-identification, cross-session association, residual sensitive attributes, and sharing of real samples, typically corresponding to PS1 to PS2. The classification criteria for the corresponding tables are thus determined: they are not only a list of references, but rather illustrate, from the perspectives of representative work and comparative dimensions, how methods in the training phase fit into the analytical framework of ``Protection Object---Lifecycle Stage---Dominant Protection-Strength Level,'' and explain their protection boundaries and residual risks through risk pathways.

User data privacy during the training phase is not a simple matter of ``preventing data leakage,'' but rather a composite attack surface resulting from the combined effects of raw input exposure, update leakage, and representation leakage. Although current methods have formed a multi-level spectrum ranging from engineering risk mitigation to formally provable protection, most work remains at the low-to-medium intensity level; simultaneously, existing mechanisms still have limited adaptability to EEG temporal characteristics, strong individual variability, and mental privacy. Therefore, how to provide more stable, transferable, and comparable tiered protection for highly sensitive user data in brain-computer interfaces while ensuring model performance remains the core issue in this field.

\begin{table}[htbp]
\centering
\footnotesize
\setlength{\tabcolsep}{3pt}
\renewcommand{\arraystretch}{1.35}
\caption{Summary of Privacy-Preserving Methods for User Data During the Training Phase}
\label{tab:3}
\begin{adjustbox}{max width=\textwidth,max totalheight=0.82\textheight}
\begin{tabularx}{\textwidth}{L{0.13\textwidth} L{0.19\textwidth} L{0.18\textwidth} L{0.32\textwidth} L{0.05\textwidth}}
\toprule
Reference & Method & \makecell[l]{Risk pathway\\mitigated} & \makecell[l]{Scenario /\\protected carrier} & PS \\
\midrule
Ju et al., 2020 \cite{ju2020_041} & Federated transfer learning & Cross-site raw-data exposure & EEG classification; raw data and identity-linked features & PS2 \\

Amato et al., 2025 \cite{amato2025_042} & Federated multicenter BCI training & Centralized sharing exposure & Multicenter BCI training; raw neural data & PS2 \\

Jia et al., 2026 \cite{jia2026_043} & SAFE: FL with local BN, adversarial training, and weight perturbation & Raw-data sharing, update leakage, adversarial misuse & EEG decoding; raw data and local features & PS2 \\

Yin et al., 2022 \cite{yin2022_046} & DP-FL for sleep staging & Membership and attribute inference & Sleep staging; EEG data and sleep labels & PS3 \\

Luo et al., 2025 \cite{luo2025_045} & DP-FL for epilepsy recognition & Membership and attribute inference & Epilepsy recognition; EEG data and disease labels & PS3 \\

Fu et al., 2024 \cite{fu2024_047} & DP-based multimodal representation learning & Representation leakage, membership inference, attribute inference & EEG representation learning; features and sensitive attributes & PS3 \\

Paul et al., 2025 \cite{paul2025_051} & Federated split learning with secure aggregation, RDP, and generative synthesis & Raw-data exposure, update leakage, representation leakage & Collaborative EEG generation/training; raw EEG and features & PS2 \\

Meng et al., 2023 \cite{meng2023_052} & Identity-unlearnable perturbation & Re-identification and cross-session linkage & EEG decoding/data sharing; identity features & PS1 \\

Meng et al., 2024 \cite{meng2024_053} & Multi-attribute privacy perturbation & Identity and attribute inference & EEG training/data release; identity, gender, experience & PS1 \\

Chen et al., 2025 \cite{chen2025_054} & A3E adversarial representation learning & Identity leakage, residual sensitive attributes & EEG decoding and cross-subject transfer; identity-related features & PS1 \\

Chen et al., 2025 \cite{chen2025_055} & PAT adversarial transfer & Cross-subject identity leakage & EEG decoding transfer; identity-related features & PS1 \\

Huang et al., 2024 \cite{huang2024_056} & Privacy-preserving multi-source domain adaptation & Domain and identity leakage & Motor-imagery EEG; domain and identity features & PS1 \\

Li et al., 2022 \cite{li2022_057} & Meta-learning for privacy-preserving source knowledge transfer & Source-transfer leakage & EEG-BCI transfer learning; source-domain features & PS1 \\

Debie et al., 2020 \cite{debie2020_058} & Privacy-preserving GAN synthesis & Real-sample sharing exposure & EEG sharing/training; raw EEG samples & PS1 \\

Pascual et al., 2021 \cite{pascual2021_059} & EpilepsyGAN synthetic data & Sharing exposure of epilepsy EEG & Epilepsy detection; EEG data and disease status & PS1 \\

Chen et al., 2025 \cite{chen2025_060} & Diffusion-based data condensation & Sample sharing, re-identification & EEG extraction/training; raw-EEG distributions & PS1 \\
\bottomrule
\end{tabularx}
\end{adjustbox}
\end{table}

\subsection{User Data Privacy in the Inference Phase}

Compared to the training phase, research on user data privacy in the inference phase is more fragmented, but its practical importance is no less significant. Major threats in this phase include the plaintext exposure of raw EEG or features during upload and edge offloading; the copying or replay of authentication templates and intermediate representations; attackers using output results and multi-round queries to infer user states; and long-term behavioral and psychological profiles formed through the accumulation of continuous interactions \cite{popescu2021_014,shokri2017_015,yan2024_021,dowlin2016_061,juvekar2018_062,pereteanu2022_063,choquettechoo2021_064,wang2022_065,wang2023_066,zhang2024_067,koppikar2016_068}. In terms of data carriers, the inference stage involves real-time inputs, local features, templates, output results, feedback logs, and long-term derivatives; in terms of risk pathways, the focus extends beyond input confidentiality to include output constraints, interface behavior, and interaction boundaries.

\subsubsection{Cryptographic Inference}

One important research direction focuses on cryptographic inference and secure inference. Its core objective is to complete model inference without the server ever coming into contact with plaintext inputs, thereby protecting the neural data uploaded by users. Works such as CryptoNets, GAZELLE, SecureML, and SecureNN have laid the technical foundation for encrypted inference, secure multi-party computation, and secure neural-network protocols \cite{mohassel2017_048,dowlin2016_061,juvekar2018_062,wagh2019_049,lindell2020_030}; in brain-computer interface scenarios, related methods have already been applied to tasks such as EEG classification, depression recognition, epilepsy detection, and brainprint recognition \cite{popescu2021_014,park2023_071,dong2023_072}. According to the classification in this review, homomorphic encryption, secure multi-party computation, and related cryptographic inference schemes belong to the PS4 cryptographic strong protection level: when corresponding cryptographic assumptions and protocol boundaries are satisfied, they first block direct visibility of plaintext inputs, critical intermediate results, or model assets, and then constrain subsequent exploitation at the protocol level. However, the engineering costs are equally significant, primarily manifested as high computational, memory, and latency overheads, as well as the requirement to modify model structures to be ``cryptography-friendly'' \cite{park2023_071,moon2024_073}. Therefore, cryptographic reasoning represents a higher level of input protection in brain-computer interfaces (BCIs), but there remains a significant gap before it can be applied to real-time closed-loop control and large-scale deployment.

The value of cryptographic inference lies in directly cutting off the server's access to plaintext neural inputs; it is the approach that comes closest to achieving strong protection during the inference phase. Its shortcomings---computational overhead, model structure constraints, and real-time performance pressures---mean that it still requires extensive engineering optimization for online BCI closed-loop control.

The security of homomorphic encryption, particularly fully homomorphic encryption, is based on standard cryptographic security assumptions. Typical objectives include ensuring that ciphertext is indistinguishable in a semantic security sense and allowing circuit computations to be performed on ciphertext without decryption. Therefore, based on the definition provided earlier, homomorphic encryption belongs to the category of strictly protected methods with formal guarantees \cite{gentry2009_029,armknecht2015_070}.

In EEG and brain-computer interface (BCI) scenarios, this approach has begun to be applied to EEG classification and disease identification tasks. For example, existing research has incorporated homomorphic encryption into the EEG classification process, enabling models to perform inference on encrypted data and thereby preventing plaintext EEG data from being directly exposed to the server \cite{popescu2021_014};

Similarly, other works have combined FHE with CNNs for applications such as depression recognition, epilepsy detection, and brainwave pattern recognition \cite{park2023_071,dong2023_072}. These studies demonstrate that privacy-preserving EEG inference has moved beyond general cryptographic frameworks and is now extending to specific neural signal tasks.

However, it must be emphasized that most of these efforts still face significant engineering bottlenecks. Existing literature has repeatedly pointed out that the primary bottleneck in homomorphic encryption inference lies in high computational and memory overhead. Consequently, it is often necessary to perform ``cryptography-friendly'' modifications to model structures and ciphertext computation processes, such as adopting activation approximation functions better suited for FHE, optimizing packing strategies, and improving matrix multiplication implementations \cite{park2023_071,moon2024_073}. Consequently, while secure inference can theoretically significantly reduce the risk of input leakage, the trade-off often manifests in increased inference latency, higher computational costs, and constraints on model architecture---which poses a particular challenge for BCI systems that emphasize real-time closed-loop control.

It is important to emphasize that user data privacy during the inference phase extends beyond input encryption. Even if raw neural inputs are strongly protected, output results, confidence distributions, and multi-round query interfaces can still be exploited to infer user states or training-related information \cite{shokri2017_015,choquettechoo2021_064}. The true challenge during the inference phase lies in simultaneously constraining both ``input visibility'' and ``output inferability.'' From a Protection Strength perspective alone, input protection schemes may have reached Level 3 or even Level 4; however, if output interface governance is lacking, the overall system's dominant Protection Strength (PS) level may still remain at a lower tier. This is also a key distinction between user data privacy during the inference phase and the training phase: the protection effectiveness is more susceptible to being undermined by interface design and long-term interaction mechanisms.

\subsubsection{Hybrid Inference Architectures}

An important approach for the inference stage involves privacy-preserving feature extraction at the edge and hybrid inference architectures. Such methods typically perform protected feature processing on raw neural signals at the user end or edge, then use the resulting intermediate representations for subsequent predictions, thereby achieving a trade-off among privacy preservation, computational overhead, and system latency. LightPyFE, proposed by Yan et al., outsources the feature extraction process to edge servers for collaborative completion via additive secret sharing and secure interaction protocols, thereby reducing opportunities for direct exposure of raw EEG data \cite{yan2024_021}. In terms of security levels, such schemes generally fall between Level 2 (engineering isolation and process constraints) and Level 3 (statistical perturbation and limited provability): they can constrain information visibility within critical processing links but often fail to achieve end-to-end confidential protection throughout the entire process and struggle to directly control inference risks at the output level. For online BCI systems that prioritize low latency and edge deployment, this type of hybrid architecture---which offers localized reinforcement and overall deployability---holds practical appeal; however, its protection boundaries must be evaluated in conjunction with subsequent inference, feedback, and log management.

The value of hybrid inference lies in its closer alignment with real-world deployments, allowing for a trade-off between local computing power, cloud-based model capabilities, and latency requirements; the risk is that intermediate representations are not inherently secure, and without additional constraints, they may still leak identity or non-target states.

\subsubsection{Template Protection Mechanisms}

In scenarios such as EEG authentication and brainprint recognition, template protection and revocable representations constitute another branch of the inference stage.

Related research protects brainprint templates in authentication systems through methods such as revocable templates, perceptual hashing, and homomorphic encryption matching, enabling their replacement or revocation after a leak, thereby preventing users from being exposed to the same immutable biometric template over the long term \cite{wang2022_065,wang2023_066,zhang2024_067,koppikar2016_068}. From the perspective of this review's framework, such work protects identity templates and authentication representations in outputs and long-term derivatives, primarily blocking exposure, replay, and cross-session correlation paths; their strength depends on whether they rely solely on template replacement and access control, or are further combined with cryptographic matching protocols. The former typically falls under PS1 to PS2, while the latter can be elevated to PS4.

The highlighted value of template protection lies in addressing the irreplaceability of long-term stable features such as brainprints, using revocable and replaceable designs to reduce the lifetime risk caused by a single leak; however, it primarily protects the authentication carrier and cannot automatically eliminate residual emotional, disease, or cognitive attributes within the template.

\subsubsection{Summary}

When viewed within a three-dimensional framework of ``Protected Object---Lifecycle Stage---Dominant Protection-Strength Level'' and in conjunction with risk pathways, research on user data privacy during the inference stage can be divided into three main lines: cryptographic inference primarily mitigates the exposure of plaintext in online inputs and falls under PS4; edge feature extraction, distributed computation, and hybrid architectures primarily reduce the leakage of raw inputs and local features, typically falling under PS2; template protection and revocable representations primarily mitigate brainprint template leakage, replay attacks, and irreversible identity exposure, typically falling under PS1--PS3. The classification here is based not on the method names themselves, but on the extent to which their dominant mechanisms can prevent the re-identification of individuals from inputs, features, or templates.

Overall, research on user data privacy during the inference stage has gradually expanded from ``verifying the feasibility of ciphertext inference'' to more specialized areas such as ``optimization for edge deployment'' and ``authentication template protection,'' but there remains a clear bias toward input confidentiality. Compared to the relatively mature collaborative training privacy frameworks in the training phase, the inference phase still has significant shortcomings regarding output inferability, continuous interaction leakage, and long-term profiling control. This implies that even if individual mechanisms achieve a high dominant protection-strength level, the system as a whole may remain at a medium-to-low strength level due to insufficient governance of output interfaces and usage processes.

\begin{table}[htbp]
\centering
\footnotesize
\setlength{\tabcolsep}{3pt}
\renewcommand{\arraystretch}{1.35}
\caption{Summary of Privacy-Preserving Methods for User Data During the Inference Phase}
\label{tab:4}
\begin{adjustbox}{max width=\textwidth,max totalheight=0.82\textheight}
\begin{tabularx}{\textwidth}{L{0.14\textwidth} L{0.21\textwidth} L{0.20\textwidth} L{0.29\textwidth} L{0.05\textwidth}}
\toprule
Reference & Method & \makecell[l]{Risk pathway\\mitigated} & \makecell[l]{Scenario /\\protected carrier} & PS \\
\midrule
Popescu et al., 2021 \cite{popescu2021_014} & HE-based EEG classification & Plaintext input exposure & EEG classification; real-time EEG input & PS4 \\

Dong et al., 2023 \cite{dong2023_072} & FHE with CNN for depression recognition & Inference-time plaintext exposure & Depression recognition; EEG input and disease status & PS4 \\

Yan et al., 2024 \cite{yan2024_021} & LightPyFE: low-leakage edge feature extraction & Raw-input exposure at feature extraction & Edge EEG processing; raw EEG and local features & PS2 \\

Wang et al., 2022 \cite{wang2022_065} & Cancelable EEG biometrics & Template leakage and replay & EEG biometrics; brainprint templates and identity privacy & PS1-PS2 \\

Wang et al., 2023 \cite{wang2023_066} & PolyCosGraph revocable brain-pattern authentication & Template leakage, replay, irreversibility & EEG biometrics; brainprint templates and identity privacy & PS1-PS2 \\

Zhang et al., 2024 \cite{zhang2024_067} & Feature homomorphic encryption for brainprint recognition & Feature exposure and template leakage & Brainprint recognition; brainprint features and identity privacy & PS4 \\

Koppikar, 2016 \cite{koppikar2016_068} & Perceptual hashing for EEG authentication & Template reuse and leakage & EEG authentication; brainprint templates and identity privacy & PS1-PS2 \\
\bottomrule
\end{tabularx}
\end{adjustbox}
\end{table}

\section{Model Privacy}

\label{sec:model-privacy}

As used in this review, ``model privacy'' refers to the protection of BCI model parameters, gradients, embedding spaces, decision boundaries, inference interfaces, and other model byproducts. It ensures that the neural data, attributes, or state information of the individuals used for training are not disclosed, recovered, or inferred through the training or service processes, nor are they illegally stolen or misused. Unlike user data privacy, model privacy is not a separate concern detached from the data; rather, it represents an extended risk that arises once data enters the model. As a vehicle for processing, compressing, storing, and reusing neural data, the model may transform privacy risks originally attached to the data into algorithmic risks that can be deployed, invoked, replicated, and exploited. Particularly in BCI scenarios, model leakage directly undermines data privacy, as parameters, embeddings, decision boundaries, and interface outputs may encode the neural features, disease patterns, emotional states, and cognitive differences of the training subjects.

\subsection{Model Privacy During the Training Phase}

Research on model privacy during the training phase began later than research on user data privacy, but its importance is no less significant. Core threats at this stage include membership inference, gradient inversion, sample reconstruction, attribute inference, and the exposure of model mechanisms resulting from anomalous updates and malicious aggregation \cite{xia2023_004,shokri2017_015,nasr2019_034,zhu2019_035,geiping2020_036,ganju2018_037,melis2018_038,zhang2024_039,r2026_040}. Even if raw EEG data is not directly shared, parameter updates, gradients, and intermediate representations exchanged during training may still be used to infer training samples, labels, or individual attributes. For BCIs, this implies that model assets are not neutral computational byproducts but may serve as compressed carriers of neurologically sensitive information.

From the perspective of attack mechanisms, existing general research primarily focuses on three categories of risks. The first is instance inference, which involves determining whether a specific data point was used in training by analyzing differences in model behavior \cite{shokri2017_015,nasr2019_034}; the second is gradient inversion and sample reconstruction, which involves approximating the original input or labels from shared gradients \cite{zhu2019_035,geiping2020_036}; the third is attribute or property inference, which involves inferring the statistical properties of the training set and the implicit label distribution from the updated or trained model \cite{ganju2018_037,melis2018_038}. For brain-computer interfaces, these attacks are particularly sensitive because EEG signals themselves may carry highly sensitive information such as identity, medical conditions, emotions, and cognitive abilities. Once model updates become reversible, what is leaked is no longer merely the training mechanism, but potentially the user's neural characteristics themselves.

\subsubsection{Differential Privacy}

Differential privacy is a representative statistical perturbation and provable method for model privacy during the training phase. By introducing noise into gradients, parameter updates, or intermediate variables, it imposes quantifiable constraints on the influence of individual training samples on model outputs, thereby mitigating membership inference and partial inversion attacks \cite{abadi2016_044}. In the classification used in this review, differential privacy primarily corresponds to PS3 rather than the highest level, PS4. This is because the method controls statistical leakage boundaries and often faces significant trade-offs between privacy budget, recognition accuracy, and model performance in BCI scenarios characterized by high dimensionality, low signal-to-noise ratios, and strong individual variability.

The value of differential privacy in model privacy lies in transforming the question of ``whether the model remembers a specific neural sample'' into a measurable problem; its limitation is that noise can impair the BCI model's ability to learn weak signals and individual variations.

\subsubsection{Secure Aggregation}

Secure aggregation and secure collaborative training mechanisms primarily constrain the visibility of individual client updates during training.

The protocol proposed by Bonawitz et al. ensures that the server can only observe the aggregated results of updates, while being unable to access the plaintext updates of individual clients \cite{bonawitz2017_033}. In terms of security levels, such mechanisms are better classified under Level 2 (engineering isolation and process constraints) to Level 3 (statistical perturbation and restricted provable privacy) protection: they can significantly reduce the chances of direct exposure in a single update round, but do not provide a global leakage upper bound in the sense of differential privacy. For multi-center collaborative modeling and cross-institutional training in brain-computer interfaces, these methods hold strong practical value, but their protection boundaries remain limited to specific participants, specific protocols, and specific training processes.

The value of such mechanisms lies in reducing the risk of single-round updates being directly stolen or observed; however, without cross-round protection and restrictions on anomalous queries, the aggregated results may still be exploited for differential analysis or attribute recovery.

\subsubsection{Federated Learning}

Beyond differential privacy and secure aggregation, efforts in federated learning---such as reputation mechanisms, security auditing, and malicious client blocking---are more focused on enhancing training security rather than strict model privacy protection \cite{zhang2024_039,r2026_040}. By identifying anomalous updates and reducing the risks of poisoning and backdoors, they indirectly limit the opportunities for attackers to exploit model updates. Classified by protection strength, this approach primarily corresponds to Level 1 (basic exposure control) and Level 2 (engineering isolation and process constraints). Its role is not to directly provide provable privacy boundaries, but rather to enhance the overall robustness and controllability of the training process.

Its value lies in maintaining the trustworthiness of the collaborative training process and preventing malicious updates from amplifying the risk of model leakage or contamination; however, it primarily addresses training security and robustness issues and cannot replace privacy proofs targeting neural data memory and inversion.

\subsubsection{Summary}

When viewed within the three-dimensional framework of ``Protected Object---Lifecycle Stage---Dominant Protection-Strength Level'' and in conjunction with risk pathways, model privacy research during the training phase primarily focuses on membership inference, gradient inversion, and attribute inference. These attacks demonstrate that model parameters, gradient updates, and intermediate representations are not neutral byproducts of training but may carry information about training samples and sensitive attributes. Accordingly, differential privacy primarily corresponds to PS3; secure aggregation and trusted execution environment (TEE)-based mechanisms mainly fall under PS2; while reputation mechanisms, secure auditing, and malicious client blocking primarily fall under PS1--PS2. If secure multi-party computation or cryptographic training protocols are introduced, PS4 can be achieved in specific stages.

Model privacy during the training phase in the field of brain-computer interfaces remains, overall, at a stage characterized by ``extensive adaptation of general-purpose methods but insufficient scenario-specific theory.'' Although existing mechanisms cover a multi-level spectrum ranging from procedural constraints to formal guarantees, specialized designs targeting EEG temporal features, inter-subject variability, small-sample training, and high-order neural attribute compression mechanisms remain relatively limited. More importantly, current research has yet to fully address a problem unique to brain-computer interfaces: exactly which neural attributes would be leaked if model updates were reverse-engineered, and how such leakage should be evaluated within the prevailing PS-level framework.

Therefore, model privacy issues during the training phase cannot be simply viewed as a replication of general federated learning security problems in BCI; rather, they should be understood as a problem of compressing and embedding neurally sensitive information within model mechanisms. The key to future research lies not only in continuing to draw upon general tools such as differential privacy and secure aggregation, but also in elucidating exactly which neural attributes are embedded in model assets, how these attributes can be inverted or inferred, and what level of protection corresponding mechanisms can achieve within the PS framework.

\begin{table}[htbp]
\centering
\footnotesize
\setlength{\tabcolsep}{3pt}
\renewcommand{\arraystretch}{1.35}
\caption{Summary of Privacy-Preserving Methods for Model Assets During the Training Phase}
\label{tab:5}
\begin{adjustbox}{max width=\textwidth,max totalheight=0.82\textheight}
\begin{tabularx}{\textwidth}{L{0.13\textwidth} L{0.21\textwidth} L{0.19\textwidth} L{0.31\textwidth} L{0.05\textwidth}}
\toprule
Reference & Method & \makecell[l]{Risk pathway\\mitigated} & \makecell[l]{Scenario /\\protected carrier} & PS \\
\midrule
Jia et al., 2026 \cite{jia2026_043} & SAFE: FL with local BN, adversarial training, and weight perturbation & Anomalous updates, partial leakage & EEG decoding; local parameters, updates, distribution traces & PS2 \\

Zhang et al., 2024 \cite{zhang2024_039} & Reputation-based federated defense & Malicious clients, abnormal updates, poisoning & EEG classification; model updates and training integrity & PS2 \\

R. S. S. et al., 2026 \cite{r2026_040} & Security audit and malicious-client blocking & Abnormal updates, malicious manipulation & Federated seizure detection; updates, logs, collaborative state & PS1 \\

Yin et al., 2022 \cite{yin2022_046} & DP-FL for sleep staging & Membership inference, partial inversion & Sleep staging; sample traces in model updates & PS3 \\

Luo et al., 2025 \cite{luo2025_045} & DP-FL for epilepsy recognition & Membership inference, partial inversion & Epilepsy recognition; disease-linked sample traces in updates & PS3 \\

Fu et al., 2024 \cite{fu2024_047} & DP-based representation learning & Representation leakage, membership inference, attribute inference & EEG representation learning; representations and features & PS3 \\

Paul et al., 2025 \cite{paul2025_051} & Federated split learning with secure aggregation and RDP & Single-client update visibility, representation leakage & Collaborative EEG generation/training; updates, representations, distributions & PS2 \\
\bottomrule
\end{tabularx}
\end{adjustbox}
\end{table}

\subsection{Model Privacy During the Inference Phase}

Compared to the training phase, model privacy during the inference phase primarily focuses on the risks of model extraction, inversion, and misuse during deployment and service. In BCI scenarios, the primary threats at this stage include model extraction and functional replication under black-box queries, confidence-driven model inversion and attribute inference, illegal redistribution or redeployment of models, as well as parameter theft, memory snooping, and execution tampering on the runtime environment side \cite{tramr2016_074,fredrikson2015_075,mehnaz2022_076,juuti2019_077,xu2023_078,abdelaziz2025_079,wang2024_080,narra2019_081,sun2024_082,papafragkaki2025_083,ding2024_084,zhao2025_085}. The consequences of these risks extend beyond damage to model intellectual property; they may also lead to the indirect leakage of disease patterns, emotional characteristics, and cognitive differences encoded in the training data.

Unlike the training phase, where model privacy primarily revolves around gradient and parameter updates, the core risks in the inference phase are more evident in model extraction and model inversion triggered by interface exposure. Tram\`{e}r et al. demonstrated that attackers can efficiently approximate and replicate the functionality of a target model via a Prediction API \cite{tramr2016_074}, while Fredrikson et al. showed that confidence information can be exploited to infer sensitive features related to the training data \cite{fredrikson2015_075}. Subsequent research further indicates that even with only black-box access, attackers may recover implicit attributes from training samples through model inversion and attribute inference \cite{mehnaz2022_076}. For brain-computer interfaces, this implies that model leakage during the inference phase is not only an intellectual property risk but may also serve as a gateway for the indirect exposure of neural features and training distributions.

\subsubsection{Interface-Level Protection}

To address the aforementioned risks, existing research has primarily focused on interface-level protection approaches, including mechanisms such as query throttling, output pruning, confidence suppression, and access auditing. The core idea is to reduce the success rate of model extraction and model inversion attacks by limiting the amount of information an attacker can obtain during black-box interactions \cite{tramr2016_074,fredrikson2015_075,juuti2019_077}. In terms of protection strength, such schemes typically fall within the scope of Level 1 (basic exposure control) to Level 2 (engineering isolation and process constraints): they can significantly increase the cost of an attack, but rarely directly alter the replicability of the model itself, nor do they provide a unified formalized leakage boundary. For brain-computer interface systems, the value of these methods lies primarily in blocking persistent query-based attacks, but their residual risk still depends on output design, audit strength, and anomaly detection capabilities.

The value of interface-level protection lies in its low deployment cost and its ability to directly reduce the attacker's information space; its limitation is that it relies more on service policies and anomaly detection, making it difficult to provide strong guarantees against high-patience, low-frequency, or distributed queries.

\subsubsection{Model Ownership Protection}

Model ownership protection represents another critical approach. Its objective is not to block every single unauthorized query, but rather to ensure that the original ownership can still be proven even after the model has been copied, misappropriated, or redeployed. EEG model watermarking and fingerprinting techniques have begun to emerge, such as the watermarking mechanism by Xu et al., the cryptographic authentication-based watermarking framework by Abdelaziz et al., and the diffusion model-based fingerprint construction method by Wang et al. \cite{xu2023_078,abdelaziz2025_079,wang2024_080}. In terms of protection strength, this approach primarily falls within Levels 1 (basic exposure control) to Level 2 (engineering isolation and process constraints): it emphasizes post-incident tracking and attribution verification rather than preemptively preventing model leakage itself. Its particular significance in brain-computer interfaces is that, once a model is copied or back-engineered, the consequences of leakage extend beyond the theft of commercial models to potentially include the spillover of neural attributes and population characteristics from the training distribution.

The value of model watermarking and fingerprinting lies in providing evidence for accountability after model theft, but it does not directly prevent the leakage of training sample information within the model; therefore, it must be used in conjunction with anti-reverse engineering protection, interface control, and data privacy mechanisms.

Consequently, model privacy during the inference stage in brain-computer interfaces cannot be simply understood as a general issue of model intellectual property. For general machine learning services, model extraction primarily threatens the commercial model itself; however, in BCI scenarios, once a model is copied, reverse-engineered, or subjected to prolonged queries, the risks extend further to the disease patterns, emotional characteristics, and cognitive differences encoded in the training data. This is also the key distinction between model privacy during the inference phase of brain-computer interfaces and that of general MLaaS models: the consequences are not only that ``the model is stolen,'' but may also entail the indirect leakage of highly sensitive neural attributes.

\subsubsection{Trusted Execution Environment}

The Trusted Execution Environment (TEE) provides another deployment-level protection pathway. Its core concept is to execute inference within a hardware-isolated environment, thereby safeguarding model parameter confidentiality while ensuring runtime integrity \cite{narra2019_081}. Relevant research has mitigated enclave memory and performance bottlenecks through model partitioning, computational offloading, and runtime optimization, and hybrid designs combining homomorphic encryption with SGX have also been explored for privacy-preserving neural-network inference, indicating that the feasibility of TEE in practical inference services is improving \cite{xiao2021_069,sun2024_082,papafragkaki2025_083,ding2024_084}. In terms of security strength classification, TEEs typically fall within the range of Level 2 (engineering isolation and process constraints) to Level 3 (statistical perturbation and limited provability): they can significantly limit runtime parameter theft and memory snooping, but their effectiveness depends on hardware trust boundaries, system implementation, and deployment conditions. For brain-computer interfaces, such mechanisms are better suited as a model deployment-level protection approach rather than as a mature, dedicated solution.

The value of TEE lies in providing an engineering-deployable isolation boundary for models and inputs, making it suitable for low-latency inference scenarios; its limitation is that security depends on hardware roots of trust, side-channel protection, and the quality of runtime implementation.

\subsubsection{Summary}

When viewed within the three-dimensional framework of ``Protected Object---Lifecycle Stage---Dominant Protection-Strength Level'' and in conjunction with risk pathways, model privacy research in the inference stage has already established a clear division of functions: Interface-level protection primarily mitigates model extraction and model inversion under black-box queries, typically corresponding to PS1--PS2; Model watermarking and fingerprinting are primarily used for attribution verification after the model is copied, reposted, or redeployed, typically corresponding to PS1; trusted execution environments primarily mitigate runtime parameter theft, memory snooping, and unauthorized access, typically corresponding to PS2; cryptographic inference and secure protocols correspond to PS4 if they can complete inference without exposing plaintext inputs or critical model assets.

\begin{table}[htbp]
\centering
\footnotesize
\setlength{\tabcolsep}{3pt}
\renewcommand{\arraystretch}{1.35}
\caption{Summary of Privacy-Preserving Methods for Model Assets During the Inference Phase}
\label{tab:6}
\begin{adjustbox}{max width=\textwidth,max totalheight=0.82\textheight}
\begin{tabularx}{\textwidth}{L{0.13\textwidth} L{0.21\textwidth} L{0.19\textwidth} L{0.31\textwidth} L{0.05\textwidth}}
\toprule
Reference & Method & \makecell[l]{Risk pathway\\mitigated} & \makecell[l]{Scenario /\\protected carrier} & PS \\
\midrule
Xu et al., 2023 \cite{xu2023_078} & EEG model watermarking & Model theft and redeployment & EEG model IP protection; ownership and parameters & PS1 \\

Abdelaziz et al., 2025 \cite{abdelaziz2025_079} & EEG neural-network watermarking / authentication & Model theft, post-tuning redistribution & EEG model IP protection; ownership and integrity & PS1 \\

Wang et al., 2024 \cite{wang2024_080} & Diffusion-based fingerprinting & Copied-model tracing & EEG model IP protection; ownership and identity tags & PS1 \\

Papafragkaki et al., 2025 \cite{papafragkaki2025_083} & TEE deployment optimization with InferONNX & Runtime theft, memory snooping, unauthorized access & Secure inference service; parameters, runtime state, partial inputs & PS2 \\
\bottomrule
\end{tabularx}
\end{adjustbox}
\end{table}

\section{Open Discussion}

\label{sec:open-discussion}

Mental privacy and neuroethical risks are best treated as open issues. They are not protection objects on the same technical plane as user data privacy or model privacy; rather, they emerge from the combined effects of neural data, model representations, output feedback, and long-term interaction trajectories. Ienca and Andorno discuss mental privacy, cognitive liberty, and mental integrity as candidate rights in the era of neurotechnology, and Magee, Ienca, and Farahany further highlight mental-privacy risks arising from cognitive biometrics beyond neural data \cite{magee2024_002,ienca2017_010}. The central question is therefore not whether a specific file has leaked, but whether BCI systems can reveal or reconstruct psychological content that exceeds task necessity.

The first issue that warrants attention is the technical boundary for inferring task-irrelevant sensitive attributes. Many BCI tasks inherently require the identification of motor intentions, emotions, fatigue, or cognitive load; thus, the decoding of mental states cannot simply be prohibited. What truly needs to be defined is the boundary between task-necessary information and information that crosses that boundary: emotion recognition should not simultaneously reveal identity and predispositions to illness; rehabilitation control should not accumulate cognitive weaknesses over the long term; and identity authentication should not form transferable psychological profiles. Future work needs to establish evaluations of task-irrelevant sensitive attributes, interpretable analysis of representations, and metrics for task-irrelevant information leakage specifically for BCIs.

The second issue concerns ethical risks at the population level. The leakage of a single user's neural data is already sufficiently sensitive, but large-scale neural datasets and cross-platform models will further give rise to risks such as population profiling, mass manipulation, discrimination based on cognitive differences, and behavioral prediction. If future consumer-grade, medical, and workplace BCI systems continue to accumulate population-level neural representations, the risk will no longer be limited to the identification of a single individual but may evolve into the misuse of neural data resources belonging to specific groups or even all of humanity.

The third issue concerns the ethical boundaries of thought reconstruction and the reconstruction of mental content. Current technology does not yet allow for the stable reading of complete thoughts, but advances such as continuous language semantic reconstruction, speech perception decoding, and low-density EEG image reconstruction have demonstrated that neural data and its model-based representations are approaching issues of mental integrity, freedom of thought, and personal identity---areas that traditional privacy theories struggle to cover \cite{szoszkiewicz2025_009,tang2023_092,dfossez2023_093,guenther2024_094,ienca2017_010}. Governance of these risks cannot wait until the technology is fully mature.

The fourth issue is that technical constraints and legal initiatives must proceed in parallel. Technologically, system design must incorporate decoupling of non-target information, collection of the minimum necessary data, confidential processing, source protection, interface auditing, and long-term profiling suppression. Legally and in terms of governance, it is necessary to clarify the special protective status of neural data, restrictions on its use, limitations on secondary training, restrictions on cross-scenario reuse, and the boundaries of liability for overreaching inferences. Recent discussions of neurotechnology governance similarly emphasize that rights-based recommendations must be translated into concrete protections for vulnerable users and deployment contexts \cite{bublitz2025_090}. Relying solely on user consent is insufficient to address the long-term, latent, and inferable risks in BCIs.

The fifth issue is whether encrypted processing can become a long-term infrastructure. Technologies such as homomorphic encryption, secure multi-party computation, and trusted execution environments provide a path toward ``using data while keeping it invisible.'' Especially in cloud-based inference, multi-center training, and cross-institutional research, encrypted processing may become one of the ultimate research directions: systems no longer rely on obtaining plaintext neural data but instead complete training and inference under cryptographic or hardware isolation constraints. The bottlenecks remain real-time performance, model complexity, energy consumption, and edge deployment capabilities.

The sixth issue is whether protection mechanisms can be designed from the very foundations of neuroscience. Unlike general data encryption, BCIs can leverage stimulation frequency, neural response patterns, the time-frequency structure of brain signals, and personalized encoding strategies to establish protection at the acquisition device or interaction front end. Work such as space-time-coding metasurfaces demonstrates that stimulation coding, brain signal induction mechanisms, and physical-layer encryption can collectively form a source-level protection pathway \cite{xiao2025_098}. If such methods are further combined with homomorphic encryption, verifiable computation, or secure hardware, they may give rise to a new research direction characterized by ``neural mechanism constraints + cryptographic computation.''

Future BCI privacy protection should shift from point-level data protection to system-level governance: reducing unnecessary neural-information exposure in data-flow design, limiting the learning and retention of task-irrelevant sensitive attributes in model objectives, and suppressing the cumulative amplification of mental content through interface outputs and long-term feedback. Only when restrictions on task-irrelevant inference become a core design principle can BCI privacy protection move from leak prevention to boundary protection.

\section{Conclusion and Outlook}

\label{sec:conclusion-and-outlook}

\subsection{Summary of the Current State of Research}

This review addresses BCI privacy protection through an analytical logic centered on data as the source and models as the carrier. It reorganizes existing work using a three-dimensional framework of protection object, lifecycle stage, and dominant protection-strength level. Rather than listing methods only by technical category, the framework emphasizes the source-carrier relationship between user data privacy and model privacy and compares how different methods block exposure, association, inference, reconstruction, and inversion pathways. Mental privacy and neuroethical risks are discussed separately as open issues instead of being treated as parallel dimensions in the basic taxonomy.

A comprehensive review of both user data privacy and model privacy reveals that current research has already established a rich spectrum of methods. Regarding user data privacy, methods such as federated learning, edge processing, template protection, generative surrogates, secure inference, and revocable representations act upon raw signals, intermediate representations, templates, and output derivatives, respectively. Regarding model privacy, membership inference, gradient inversion, attribute inference, model extraction, and model inversion reveal the privacy-bearing attributes of model assets, while differential privacy, secure aggregation, interface protection, TEE, watermarking, and cryptographic protocols constitute the primary protection pathways.

In terms of the dominant protection-strength level distribution, current literature primarily focuses on PS1 and PS2 methods. PS3 methods are mainly concentrated in differential privacy, perturbed training, and controlled output mechanisms, while PS4 methods are primarily centered on homomorphic encryption, secure multi-party computation, confidential training, and confidential inference. High-intensity protection remains constrained by accuracy loss, communication overhead, computational latency, deployment complexity, and BCI real-time requirements.

\subsection{Key Challenges}

\paragraph{BCI privacy protection still faces the following key challenges:}

(1) The boundaries between protection objects and data forms need to be unified. Current research often conflates raw data, intermediate representations, model assets, and output derivatives, making it difficult to compare different methods. Future work should clearly distinguish between raw acquired signals, intermediate data representations, model assets, and outputs and long-term derivatives, and should clarify the potential exposure, association, inference, reconstruction, and inversion pathways corresponding to each category of carrier.

(2) Model privacy and user data privacy are highly intertwined. Models are not only computational tools; they are more likely to serve as compressed carriers of neuro-sensitive information. If gradients, parameters, embedding spaces, and inference interfaces are exploited through reverse engineering, this not only exposes the model's mechanisms but may also allow for the indirect reconstruction of individual training subjects' identity characteristics, disease patterns, and emotional cues.

(3) There is a significant trade-off between high-strength protection and BCI usability. Methods such as differential privacy, homomorphic encryption, and secure multi-party computation provide more robust tools for BCI privacy protection. However, given the high dimensionality, low signal-to-noise ratio, and strong inter-individual variability of neural signals---coupled with the low-latency and real-time feedback requirements of many applications---the trade-offs in terms of accuracy, latency, and deployment complexity become particularly pronounced.

(4) There remains a lack of unified modeling for cross-stage risks. The acquisition, training, inference, and feedback stages are often discussed in isolation, but real risks frequently stem from the interaction and accumulation of risks across multiple stages. For example, attribute leakage during training may be further exploited through interface probing after deployment, while outputs and interaction trajectories from the inference stage may conversely enhance the identification of a user's long-term state.

(5) Computable, assessable, and verifiable protection frameworks for non-target mental inference are still lacking. While ethical and legal concerns regarding mental privacy are growing rapidly, the technical domain still lacks unified definitions, evaluation metrics, and experimental paradigms. In particular, regarding whether systems additionally learn and retain task-irrelevant mental content, existing methods often provide only partial, indirect, or task-dependent assessments.

(6) Evidence for privacy-preserving methods in multimodal neural data remains significantly insufficient. Current technical literature exhibits a bias toward EEG, and there is a lack of unified evaluation benchmarks and comparable privacy-preserving experiments regarding privacy risks in fNIRS, MEG, ECoG/sEEG, implanted signals, and multimodal fusion. Future work should establish more granular risk modeling and protection methods tailored to the invasiveness, spatio-temporal resolution, long-term deployment methods, and degree of identity binding of different modalities.

\subsection{Future Research Directions}

\paragraph{To address the above challenges, future BCI privacy protection research should proceed along the following lines:}

(1) Construct an evaluation and protection framework centered on decoupling. Future research should focus on the ability to decouple data from individuals, establishing a unified threat model and evaluation metrics that cover both user data privacy and model privacy, enabling different methods to be compared under the same PS classification.

(2) Develop high-strength protection mechanisms tailored to the characteristics of neural signals. Given the high-dimensional, time-series nature of BCI data, significant individual variability, and small sample sizes, it is necessary to design differential privacy budget controls, encryption-friendly model architectures, lightweight cryptographic inference, and secure collaborative training mechanisms that are better suited for neural signal tasks.

(3) Promote cross-stage collaborative protection. From a systems engineering perspective, future work should integrate data collection security, transmission protection, training privacy, inference control, output governance, and feedback management into a unified design. Protection at a single stage that fails to coordinate with other stages can easily be undermined by long-term interactions and cumulative outputs.

(4) Incorporate the restriction of task-irrelevant sensitive attribute inference into model design objectives. Regarding mental privacy and mental health risks, future efforts should not only passively assess leakage risks after model training is complete. Instead, constraints that limit the identifiability of task-irrelevant sensitive attributes should be proactively introduced into model objectives, such as multi-task adversarial training, causal representation learning, explainable feature selection, and attribute de-correlation mechanisms.

(5) Promote the coordinated development of technical protection, ethical governance, and clinical translation. BCI privacy issues ultimately serve not only algorithmic optimization, but clinical acceptability, social trust, and technological governance capabilities. Future work should co-design informed consent, use restrictions, boundaries for secondary use, liability allocation, and international standards with technical solutions, making privacy protection a fundamental component of BCI system design.

In summary, BCI privacy protection needs to be re-examined not only because neural data are highly sensitive, but also because models can compress, store, deploy, and expose that data in ways that transform one-time leakage into continuous inference risk. A clear analytical framework, together with explicit attention to mental privacy and neuroethical risk, can support more actionable protection strategies for large-scale BCI applications.

\bibliographystyle{elsarticle-num}
\bibliography{references}

\end{document}